\documentclass[twocolumn,english]{revtex4}
\usepackage[T1]{fontenc}
\usepackage[latin9]{inputenc}
\makeatletter
\@ifundefined{textcolor}{}
{%
 \definecolor{BLACK}{gray}{0}
 \definecolor{WHITE}{gray}{1}
 \definecolor{RED}{rgb}{1,0,0}
 \definecolor{GREEN}{rgb}{0,1,0}
 \definecolor{BLUE}{rgb}{0,0,1}
 \definecolor{CYAN}{cmyk}{1,0,0,0}
 \definecolor{MAGENTA}{cmyk}{0,1,0,0}
 \definecolor{YELLOW}{cmyk}{0,0,1,0}
}

\makeatother

\usepackage{babel}
\begin{document}

\title{Open-endedness in AI systems, cellular evolution and intellectual
discussions}

\author{Kushal Shah}
\email{kushals@iiserb.ac.in}

\affiliation{Department of Electrical Engineering and Computer Science, Indian
Institute of Science Education and Research (IISER), Bhopal - 462066,
Madhya Pradesh, India.}
\begin{abstract}
One of the biggest challenges that artificial intelligence (AI) research
is facing in recent times is to develop algorithms and systems that
are not only good at performing a specific intelligent task but also
good at learning a very diverse of skills somewhat like humans do.
In other words, the goal is to be able to mimic biological evolution
which has produced all the living species on this planet and which
seems to have no end to its creativity. The process of intellectual
discussions is also somewhat similar to biological evolution in this
regard and is responsible for many of the innovative discoveries and
inventions that scientists and engineers have made in the past. In
this paper, we present an information theoretic analogy between the
process of discussions and the molecular dynamics within a cell, showing
that there is a common process of information exchange at the heart
of these two seemingly different processes, which can perhaps help
us in building AI systems capable of open-ended innovation. We also
discuss the role of consciousness in this process and present a framework
for the development of open-ended AI systems.
\end{abstract}
\maketitle

\section{Introduction}

Artificial intelligence (AI) has come a long way since the inception
of Turing machines in the 1930s \cite{Luger,Wu,Turing1950,Silver,Srinivasan}.
Its applications now range from excelling in games like Chess and
Go, to predicting certain medical conditions with high accuracy, to
self-driving cars and many others which were earlier thought to be
the exclusive domains of the human mind. Despite this phenomenonal
success, one of the biggest limitations of artificial intelligence
algorithms is that they are mostly good at doing one specific kind
of task and cannot learn other tasks without the programmer making
significant changes in the algorithm. This problem can be termed as
a lack of open-endedness, which we see in biological evolution or
human creativity \cite{Stanley}. In order to enable AI systems to
harness multiple possibilities of growth, it is important to understand
the crux of the evolutionary process and then try to implement it
in silicon. For this to happen, it is also necessary to change our
perspective of biological evolution, which is largely centered around
survival of the fittest.

The theory of biological evolution has often been assumed to imply
that life is all about maximization of an individual's survival probabilities
\cite{Dawkins}. However, numerous evidences of collective behavior
and altruism have now forced scientists to change this view and see
evolution from a broader perspective \cite{Hamilton}. A very important
feature of biological systems is that they are deeply connected and
integrated with their surrounding environment and even with other
living organisms. Most living organisms consume other living organisms
as food. Even the human system cannot function without the bacterial
life forms in its intestines. Of course, this interaction is not always
mutually beneficial since foreign bacteria and viruses can also cause
diseases, which at times can even be fatal. And the very act of consuming
other living beings as food implies that one being must die for the
other to survive. On the surface, it looks like a state of intense
competition where each organism is trying to outcompete or even kill
the other in order to maximize its own survival. Looking a bit deeper,
we find that there is also a lot of cooperation within the same species
and even across different species (eg. gut bacteria in humans). Going
still deeper, we find that both competition and cooperation are just
different forms of \emph{interaction} between organisms of the same
or different species. Hence, in some sense, what drives biological
evolution is neither competition for survival nor cooperation, but
a very deep and intricate network of interaction, which biological
evolution is perhaps trying to explore. 

We see this process of evolution by interaction in the scientific
domain too where many scientific discoveries and inventions have come
about as a result of interaction (both competitive and cooperative).
This method of interaction or discussions can also be a very effective
method of teaching in our classrooms and is usually called the \emph{Socratic
method} \cite{Neumayr,Shah2016}. Some scientists fiercely compete
with each other for being the first ones to make a certain discover
and others cooperate to work in a collaborative way. In order to fully
understand the process of scientific discovery, we need to look at
both competition and cooperation as being just different aspects under
the larger umbrella of \emph{interaction}. 

In this paper, an information theoretic analogy is presented which
suggests that, when seen from the perspective of information exchange,
there is a deep connection between intellectual discussions and cellular
evolution. This might sound a little far fetched since we usually
associate discussions with the act of exchanging \emph{words} in a
certain language. However, at its core, discussion can also be seen
as an act of exchanging \emph{information} in a certain way, which
may not necessarily have all the properties of a natural human language
\cite{Lyons}. Finally, we will see how this process of information
exchange can be incorporated into AI systems thereby imparting them
the possibility of open-ended growth.

In the next section, the discussion method is explained from an information
theoretic perspective, which will make it easy to draw analogies with
other branches of human knowledge. The basics of cellular evolution
and its connection with intellectual discussions is presented in \ref{sec:Cellular-dialogues}.
We discuss the role of consciousness in this process in Sec. \ref{sec:Role-of-consciousness}
and then present a framework for development of open-ended AI systems
in Sec. \ref{sec:Mathematical-framework}. Finally, the paper ends
with conclusion in Sec. \ref{sec:conclusion}. 

\section{Discussions as information exchange\label{sec:Discussions-as-information}}

Among all the tools and techniques developed to analyze natural systems
over the last few centuries, information theory has proved to be one
of the most powerful ones with applications in almost all areas of
science and engineering \cite{Thomas}. This is primarily because
information theory provides a way to model the most fundamental core
of various phenomenon without worrying about the higher order complications.
Thus, it is natural to also examine intellectual discussions through
this framework. However, in order to do this effectively, it is important
to understand two salient features of intellectual discussions which
differentiate it from other forms of dialogue:
\begin{enumerate}
\item \textbf{Open-ended :} In debates, the participant(s) sole objective
is to convince the other participant(s) of their own point of view,
thereby making them close-ended. In contrast, an intellectual discussion
is truly open-ended with all participants willing to genuinely listen
to each other's point of view and as a result, evolve their own perspectives
about the topic being discussed. Due to this, the amount of uncertainty
in a intellectual discussion is much higher than that in close-ended
debates. Here, it is important to note that close-ended interactions/debates
can also be beneficial in many situations, but they serve a very different
objective from that of intellectual discussions. Some examples of
beneficial close-ended interactions are doubt clearing sessions between
a teacher and individual students, or even the Upanishadic dialogues
of ancient India \cite{Upanishad}. These kinds of interactions can
also be very illuminating, but in these, one of the participants (Guru)
is supposed to already know the answers to all/most questions relevant
to the topic being discussed. This partly applies in a classroom setting
too where the teacher relatively knows a lot about the topic under
discussion, but chooses not to give the answers directly as far as
possible. However, the primary difference is that, in a Socratic class
which proceeds through open-ended discussion, the role of the teacher
is to create an interesting enough discussion through effective moderation,
which leads the students to themselves find the answers through exchange
of information. This is directly connected to the role of a programmer
in developing AI systems. As of now, the programmer hard codes the
algorithm into the computer, which then can do only as much as the
algorithm allows it to do. This is similar to the lecturing method
of teaching where the student can know only as much as the teacher
has imparted. However, in an open-ended AI system, there will be no
such limitations and the exploration of new ideas and ways of doing
things will be truly unbounded and not restricted by the original
algorithm hard coded by the programmer.
\item \textbf{Collective Welfare :} Our modern education system inevitably
ends up playing the primary role of ordering the students along a
certain hierarchy through examination marks, which forces most students
to be primarily concerned with increasing their own personal welfare.
However, in an intellectual discussion, there are no brownie points
to be scored and individuals are seldom given too much importance.
The purpose of a intellectual discussion is to pool in all the information
available with the participants, and come up with a collective solution
to the problem, thereby promoting the spirit of team work. Also, learning
from our peers can be lot more effective than learning from those
whose who are much senior to us in age and experience.  It is again
important to note that individualistic discussions and aspirations
also have their place in society and are required in certain situations
like business negotiations. Though these discussions can also increase
learning for the student, as in the case of the Upanishadic method
mentioned above, they are not scalable. Doctoral students often learn
a lot from their supervisors through such discussions, but it cannot
work at the level of undergraduate education simply because a teacher
cannot devote so much time to individual students. The Socratic method
fills this gap by enabling the teacher to simultaneously have a discussion
with all students even in a large class size (see Michael Sandel's
video lectures on justice at http://justiceharvard.org). Here again,
there is a deep connection with AI systems in the sense that trying
to develop a particular open-ended AI system is not going to be very
fruitful. What is requried is to create a network of such systems
and allow them to interact with each other so that they can collectively
explore the landscape of all possibilities. 
\end{enumerate}
Keeping these two aspects in mind, let us now try to understand the
discussion process using concepts of information theory.

Information theory essentially deals with binary strings of 0s and
1s. We could, of course, use other symbols, but 0s and 1s are in some
sense the simplest to deal with without any loss of generality. The
primary quantity of importance in this domain is the \emph{information
content} of a given string of 0s and 1s. And the information content
(or Kolmogorov complexity) of such a binary string is defined to be
the length of the shortest computer program or description (which
is also just another binary string) which can represent the given
string \cite{Thomas}. Hence, a binary string which requires a program
with a longer length to represent itself, is considered to have a
higher information content. For example, the string ``01010101010101''
has a lower information content than ``10100101011100'' since the
former can simply be represented as ``$(01)_{7}$'', where the subscript
stands for the sequence being repeated that many times. What does
it have to do with intellectual discussions?

As mentioned above, an important aspect of these discussions is to
enable the participants to recombine available information in interesting
ways to generate new ideas and not just to store the imparted information.
This is akin to increasing the \emph{intelligence} of the participants
and not just enhancing their \emph{memory}. From an information theoretic
view, increasing the intelligence of a system requires increasing
the information content of its basic thinking process, whereas increasing
memory is merely enhancing information content of its input tape \cite{Srinivasan}.
In other words, increasing intelligence is somewhat like going from
normal programming to machine learning algorithms, whereas increasing
memory is just going from a 1TB hard disk to a 2TB one. From this
perspective, the discussion method is a two step process. In the first
step, it collects the 0s and 1s from the participants and then rearranges
them, through discussions, in an order so as to increase the information
content of the resulting string. This by itself is not enough and
in the next step, it takes different components of this string and
helps all participants in seeing the intricate connections between
them. It is this second more crucial step which enhances mundane memory
to the level of sublime understanding. And this process also helps
in appreciating the importance of interaction between different parts
of the whole. 

As mentioned above, one of the salient features of the discussion
method is that it is open-ended, which means that there is no apriori
decided binary string that the participants wish to reach. All strings
are acceptable, within reasonable bounds, as long as they increase
the overall information content of the participants' thinking process.
The discussion method also aims for collective welfare, which means
that the purpose is not to maximize the information content of one
or few participants, but that of the whole group. This is quite similar
to what happens within the cell as we will see in the next section.

\section{Cellular evolution\label{sec:Cellular-dialogues}}

The cell is a very complex entity and carries out a lot of processes
and functions. We will, however, focus only on the very basic ingredients
that go into making a cell and ignore the details. At a broad level,
a cell consists of the following components \cite{Alberts}:
\begin{enumerate}
\item \textbf{DNA :} This is a double stranded string of nucleotides which
is considered to be the information carrier of the cell. The DNA of
a cell is also called its \emph{genetic material} and plays an important
role in heredity.
\item \textbf{Proteins :} These are single stranded strings of amino acids
which fold into complex structures and are the primary work horses
of a cell. It is the proteins that help in carrying out most of the
important cellular functions. Proteins can also bind to different
locations along the DNA, thereby regulating its various functions.
\item \textbf{Cell membrane :} This is the outer boundary of the cell which
protects the contents inside and gives the cell a sense of cohesiveness.
\item \textbf{Mitochondria :} This is the energy house of the cell and generates
the ATP molecules that provide the energy for carrying out various
functions of the cell.
\end{enumerate}
The DNA sequence encodes the information of the cell and the role
of evolution is in some sense to increase the information content
of this sequence. For this to happen, it is important for the DNA
sequence to be flexible and be prone to changes known as \emph{mutations}.
However, all mutations are not good for the cell's well-being and
it is important for the cell to have a mechanism to retain only those
mutations which are beneficial. This process is known as \emph{natural
selection }\cite{Dawkins}. Mutations which lead in lower functionality
are finally weeded out and those which are beneficial are retained.
The important point is that there is no way for a cell to know apriori
whether a mutation is good or bad unless it actually goes through
that mutation and experiences the resulting effects. 

This is quite similar to what happens in a discussion group. The bits
of information that the participants give during the discussion are
like the mutations in the already available information content. The
role of the moderator is then to weed out the bits which do not increase
the overall information content. But the moderator does not do this
directly on its own, very much like the cell, where the DNA by itself
cannot weed out unwanted mutations. What happens in a cell is that
bad mutations are gradually weeded out through the \emph{process}
of natural selection rather than there being an authoritative agency
doing the decision making. Similarly, the discussion has to be moderated
in such a way that the participants themselves realize collectively
that certain bits of information are not useful and need to be weeded
out, while recognizing that other bits are useful and need to be retained.
This is the function that proteins play by helping the cell in collectively
going through the process of natural selection, by determining whether
a certain mutation leads to greater functionality or not. Thus, it
is the proteins which are collectively responsible for correcting
some of the mutations and leaving out others.

In this context, it is again important to note that the cell is not
really trying to maximize its survival probability. This is because
an entity that is very concerned about its own individual survival
will not be so open to mutations. A planet, which does not mutate
or evolve, has a much longer survival probability than a cell! What
the cell is trying to increase is its information content and its
interaction with other entities in the environment. That is why we
see that higher organisms have more and more interaction abilities
and not necessarily higher survival abilities. Bacterial life forms
have been living on this planet for many more millennia than humans
and will most probably continue to live even after humans go extinct.
Many other organisms also have a much longer life span than humans. 

The cell membrane also plays an important role by providing a protective
environment to the cellular contents so that they can peacefully interact
with each other. Every group needs some kind of a boundary for it
to grow. That is perhaps why we have groups of all scales and sizes
in human societies ranging from small families to large nations and
then to the even larger global community. A group boundary at all
levels plays an important role of nurturing the individuals within.
Similarly, every class or human comunity also needs a certain boundary
so that the individuals within it can interact with each other in
a consistent and cohesive manner. If the participants of a discussion
group keep changing very frequently, it is unlikely to lead to a sustained
growth of knowledge. However, some amount of change at certain periods
of time is also necessary to allow new ideas to flow into the system.
That is a delicate balance that needs to be achieved.

Every group also needs individuals who are its energy providers, like
the mitochondria of a cell. A group where all individuals are the
same is perhaps not very exciting. We need some who are good listeners,
some who are good speakers, some who are good thinkers and some who
just have too much energy. All these various kinds of individuals
are required to provide a rich experience during intellectual discussions.

\section{Role of consciousness in open-endedness\label{sec:Role-of-consciousness}}

The process of intellectual discussions and cellular evolution that
we have discussed in Sec. \ref{sec:Discussions-as-information} and
\ref{sec:Cellular-dialogues} respectively, are both regarding living
organisms. We have presented a unifying framework using which the
highest cognitive functions of human beings can be seen in the same
light as evolution of the simplest bacteria. Hence, the process of
evolution is not just something which leads to develop various organs
and organisms, but also guides them in their various functions. The
natural question to ask now is whether a similar open-ended evolutionary
process can take place in a non-living system. Despite several attempts
so far, that has not really been the case. Scientists and engineers
have been able to build AI systems with very interesting capabilities,
but all such systems are very far from being truly open-ended \cite{Srinivasan,Stanley}.
There could be two possibilities. One is that there is a fundamental
difference between living and non-living entities, which can perhaps
never be bridged. It is now well accepted that it is very much possible
for AI systems to display human-like behaviour without having any
similarity with the cognitive process used by humans \cite{Srinivasan,Searle}.
Another is that there are no such water-tight compartments and it
is just a matter of time before we find the right algorithms which
can lead to development of open-ended AI systems. If the first possibility
is true, one can then immediately invoke the concept of consciousness,
which only living entities are thought to possess and say that it
is this which is the main cause of open-endedness seen in biological
evolution. If it is the second possibility which is true, even then
there is currently a wide gap between the living and non-living entities
and a quantum jump would be required to bridge the gap. Simply developing
smarter algorithms and faster processers is not really going to make
AI systems show open-ended behaviour. And so, even in this case, some
consideration of the properties and effects of consciousness becomes
important and one cannot rule out its role in open-endedness. In this
paper, we assume that it is the second possibility which is true.

So there are two questions that need to be answered is this regard.
What is the role of consciousness in imparting open-endedness to biological
evolution? And, how can we make non-living AI systems mimic this process?
\emph{Without getting into any of the subtle aspects of consciousness,
we propose that the primary role of consciousness is to provide a
subjective evaluation of the information content of a given string}.
In Sec. \ref{sec:Discussions-as-information}, we have explained that
the concept of Kolmogorov complexity measures information content
of a string as the length of the shortest string which can be used
to represent it. Though this is very useful from an algorithmic perspective,
it is severely limiting from the perspective of human cognition and
open-endedness. The same string of letters can mean very different
things to different individuals. A cell can also carry out various
different functions using the same set of proteins. And it is this
subjective measure of information content that drives evolution and
cognition in different directions and helps in exploring the landscape
of creative possibilities in an open-ended way. A sparrow evolved
to fly in air and an elephant evolved to walk on land mainly because
both perceive the sky and space very differently. Different scientists
and engineers discover and invent different things because they perceive
the world around them in fundamentally different ways. Hence, in some
sense, we can say that biological organisms give more weightage to
non-veridical perceptions which maximize utility rather than veridical
perceptions of reality \cite{Hoffman}.

Taking a hint from \cite{Hoffman}, one can perhaps say that the primary
limitation of AI systems is that they are too tied up with a veridical
representation of objective reality. In order to achieve open-endedness
in these systems, the first would be to allow these systems to develop
their own subjective way of perceiving things. In some sense, we need
to allow them to have their own subjective way of evaluating information
content of a given string. For this to happen and lead to open-endedness,
it is very important for the AI system to try to maximize interaction
with its surroundings in some way. A system which is geared towards
highest survival probability will not go towards open-endedness for
two important reasons. Firstly, there is lot of danger in open-ended
exploration and it may drive the organism towards extinction. Secondly,
as we mentioned earlier, bacterial life forms can survive much longer
than the human species, and hence, if survival is the main goal, there
is no motivation for open-ended evolution. Even an inanimate object
survives much longer than any living entity. 

\section{Framework for open-ended AI systems \label{sec:Mathematical-framework}}

In this section, we will present a framework for the development of
open-ended AI systems based on the ideas above. 

Firstly, we need to start working with a large interacting collection
of such AI systems instead of individuals. Even human thought tends
to stagnate in the absence of intellectual interaction and currently
there is no evidence to show that a single bacteria can evolve in
the absence of interaction with its environment. This is because each
individual being operates by a certain specific set of rules and so
can generate only a limited amount of variety. It is interaction with
other beings that leads to an update in the rules and thereby, more
novelty in the thought process or in biological evolution. It is also
important that all the AI systems in a given collection are designed
for this capability of open-ended evolution, and not just one single
individual. Evolution of one individual is heavily dependent on the
evolution of all other individuals in the group!

Secondly, various individuals in this collection should be free to
change their own rules independent of others. This follows directly
from the above point. If all entities in the collection operate by
the same rules, then there is a limit to the amount of novelty they
can produce. The very purpose of interaction is to allow different
entities in the collection to be able to independently evolve its
own set of rules, which will then in turn influence the rules of other
entities in that collection. 

Thirdly, there should also be mutual dependence in some way since
there has to be a motivation for interaction. Humans tend to be influenced
by others mainly when there is some gain to be achieved or loss to
be avoided. Evolution of simpler life forms also tends to be more
influenced by factors that are directly related to its own functioning.
Since AI systems mainly run on electricity, some of the individuals
in this collection could be generators of electric power (solar, wind,
etc.), which other AI systems can then use to run themselves. Other
ways of introducing dependence is to make some AI systems more efficient
at collecting various kinds of data and other AI systems more efficient
at processing them. This way the efficient processors will depend
on the efficient collectors. 

Fourthly, and perhaps most importantly, we need to enable AI systems
to have a subjective evaluation of the information content of each
string/data they encounter. As stated in Sec. \ref{sec:Role-of-consciousness},
this is in some trying to mimic one of the roles of consciousness
in these systems. It is important to note here is that this capability
of subjective evaluation will not necessarily make the AI system conscious
in the same way a human is. It will only enable the AI system to mimic
a certain property of consciousness with the goal of open-endedness.

So far we have discussed various ways in which AI systems can mimic
the process of intellectual discussions and biological evolution in
order to develop capabilities for open-ended innovation. However,
there is a significant way in which AI systems are different from
biological organisms and that might actually pose a significant obstacle
in this process. And this difference lies in the fact that AI systems
have much faster processing capabilities and don't get tired of doing
the same thing again and again. A classic example is the AI algorithm
that mastered the game of Go by repeatedly playing against itself
\cite{Silver}. The reason this is an obstacle is that if a system
becomes too efficient in one particular domain, it tends to stick
to it without trying to explore other capabilities. This happens in
humans too who tend to keep doing that one thing they are good at
without exploring other paths. However, the problem does not effect
humans and other biological organisms so much since only a few of
them are able to reach very high levels of expertise in one field.
Humans also tend to get bored doing the same thing after a while and
we may have to incorporate this element of boredome into the AI systems
too. We need to think of AI systems as jack of all trades and not
master of one!

\section{conclusion\label{sec:conclusion}}

In this paper, an analogy has been presented between intellectual
discussions and cellular dynamics and like all other analogies, it
is surely not a perfect one. Analogies primarily play the role of
opening our thought process to a dimension of reality we may not have
been aware of. They help in building connections between disparate
concepts which do not seem to have anything to do with each other.
This analogy between the Socratic method and cellular dynamics will
perhaps help in conveying the message that when seen from the perspective
of information theory, interaction with information exchange in all
directions is a very fundamental aspect of life and essential for
the natural evolution of all living entities. This can not only help
in developing open-ended AI systems but also in making our education
system lot more open-ended. To whatever extent possible, we need to
inculcate a habit of discussions in our classrooms, homes, offices,
academic institutions and everywhere else. Discussions and thinking,
in general, are not something we are naturally good at. As Daniel
Willingham \cite{Willingham} says, ``\emph{People are naturally
curious, but we are not naturally good thinkers; unless the cognitive
conditions are right, we will avoid thinking}''. This quality has
to inculcated in our students from a young age. And for this happen,
teachers also need to be trained in the art of effective moderation.

Apart from its pedagogic value, the above ideas may also help in enhancing
our understanding of basic cell biology. Education and research is
now becoming increasingly inter-disciplinary, which makes it important
for students and teachers of all scientific fields to have a better
appreciation for biological concepts, and vice-versa. For this to
happen, biology needs to be presented in a conceptual framework that
other scientists can relate to. Information theory can provide a powerful
bridge for this purpose and might help in linking physics with biology
in the same way it bridged statistical physics and communication engineering.
It will also be interesting to build a mathematical model for the
ideas presented in this paper and see if it can predict the outcomes
of intellectual discussions, cellular evolution and open-ended AI
systems!

\end{document}